\documentclass[12pt, a4paper]{article}

\usepackage[a4paper, margin=1in]{geometry} 
\usepackage{titlesec} 
\usepackage{authblk} 
\usepackage{graphicx} 
\usepackage{hyperref} 
\usepackage{setspace} 
\usepackage{tocloft} 
\usepackage{float}
\usepackage{amsmath}
\usepackage{media9}

\title{\textbf{An Integrated Approach to Robotic Object Grasping and Manipulation}}

\author{
    Owais Ahmed, 
    Hamza Ali Khan, 
    Muhammad Huzaifa, 
    Muhammad Areeb
}
\affil{\small Computer and Information Systems Engineering Department,\\
NED University of Engineering and Technology, Pakistan}

\date{August 2023}

\begin{document}

\maketitle

\begin{center}
    \textbf{Final Year Project}
\end{center}

\noindent\rule{\textwidth}{0.5pt} 

\vspace{1cm}

\renewcommand{\thesection}{\arabic{section}.}
\renewcommand{\thesubsection}{\thesection\arabic{subsection}.}

\begin{spacing}{1.2} 
\tableofcontents
\end{spacing}

\newpage

\begin{abstract}
\setlength{\parskip}{10pt} 
\textit{
The increasing demand for efficiency in warehouse operations has led to the integration of robotics, particularly in companies like Amazon. While robots have been successfully deployed for item transportation, object picking from shelves remains a significant challenge. This research presents an autonomous robotic system designed to fulfill a simulated order by efficiently selecting specific items from shelves.}

\textit{
A key feature of the proposed system is its ability to handle uncertain object positions within storage bins. The robot adapts dynamically, employing intelligent strategies to locate and retrieve items without prior knowledge of their exact placements. The system is built with a strong focus on autonomy, integrating advanced algorithms, control mechanisms, and sensor fusion to execute the entire picking process without human intervention.}

\textit{This work highlights the synergy between robotics, computer vision, and artificial intelligence in solving complex logistical challenges. The proposed system represents a step forward in autonomous object picking technology, with the potential to enhance warehouse efficiency and redefine modern automation practices.}
\end{abstract}

\newpage

\section{Introduction}
\setlength{\parskip}{10pt} 
The field of robotics has gained widespread attention as industries explore automation’s potential to enhance efficiency. Amazon, a leading e-commerce retailer, is at the forefront of this movement, hosting an annual robotics "picking" challenge to improve its order fulfillment processes. Teams develop sophisticated robotics hardware and software capable of recognizing, grasping, and relocating items, tackling the complexities of varying shapes, materials, and environments.

This initiative aligns with Amazon's goal of optimizing operations by improving speed, accuracy, and scalability while allowing human workers to focus on higher value tasks. The challenge encompasses technological innovation and process integration, involving conveyor systems, advanced algorithms, and real-world fulfillment setups.

Despite attracting global experts and widespread participation, the challenge highlights the ongoing complexity of automating intricate tasks. As Amazon continues to invest in robotics, this competition remains a pivotal endeavor fostering collaboration and innovation while advancing the broader field of robotics.

A key part of Amazon’s strategy focuses on the complex challenge of fulfilling customer orders. In the fast paced world of e-commerce, Amazon’s commitment to swift and precise picking and packing processes reflects its dedication to meeting customer expectations and ensuring satisfaction. By leveraging robotics technology, Amazon aims to enhance efficiency and scalability while enabling human workers to focus on more complex, value driven responsibilities.

\subsection{Motivation and Need}
The increasing importance of warehouse automation and efficient order fulfillment is a key aspect of today's e-commerce environment. As online shopping grows, the need for systems that can quickly and accurately handle large volumes of orders becomes more critical. Combining robotics and machine learning in warehouse logistics can significantly improve the speed and efficiency of order processing, leading to cost savings and higher customer satisfaction.

A crucial challenge in robotics is the complex task of grasping objects due to the variety of shapes, sizes, and textures of objects. Developing a system that can determine the precise position and orientation (6D pose) of objects and perform autonomous grasping not only enhances warehouse operations but also advances the field of robotics.

This project represents an integration of advanced technologies, including computer vision, machine learning, and robotics. By using two monocular cameras to determine the 6D poses of objects and training a machine learning model to predict optimal grasping strategies, the goal is to create a flexible system capable of handling a wide range of objects.

Recent advancements in this area include Amazon's development of robots like "Robin" and "Sparrow," which use artificial intelligence to handle numerous items and streamline order fulfillment, potentially saving significant costs annually. Additionally, research combining deep learning and 3D vision techniques has led to faster and more accurate 6D pose estimation for object grasping, showing promising results in real-world applications. These developments highlight the potential of integrating robotics and machine learning to revolutionize warehouse automation and order fulfillment processes.

\begin{figure}[h]
    \centering
    \includegraphics[width=0.8\linewidth]{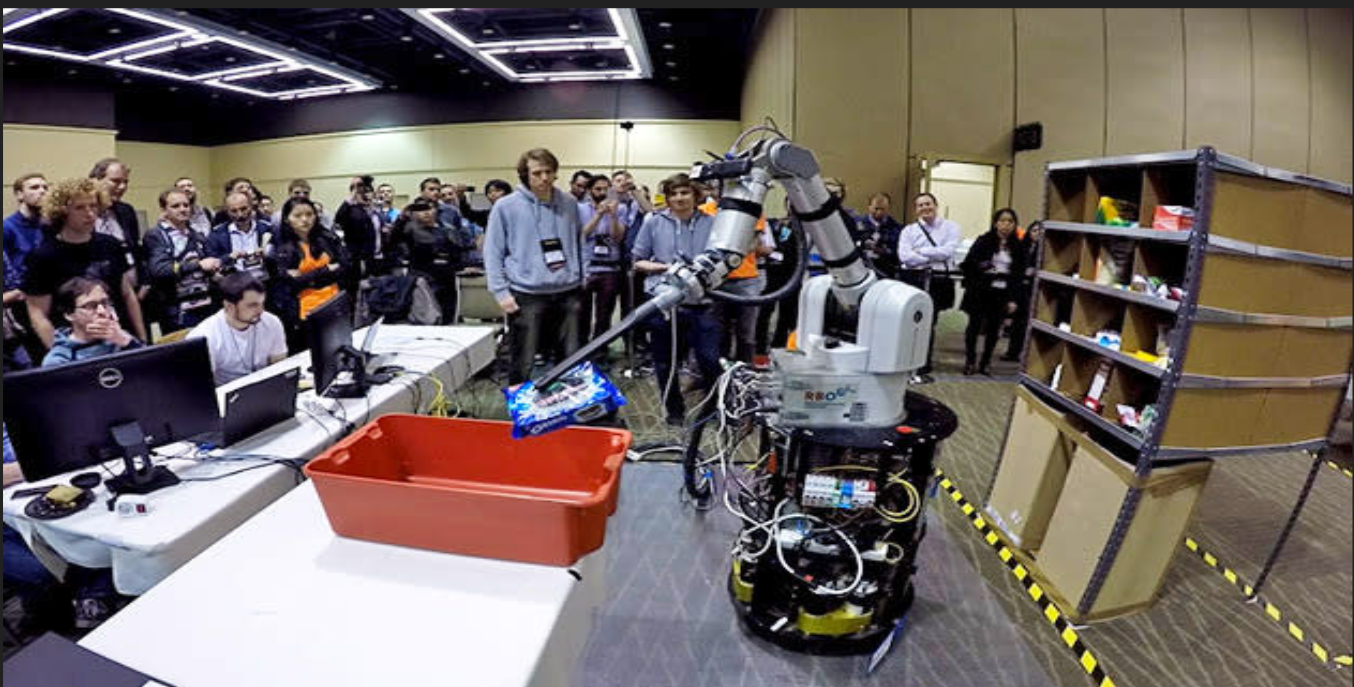}
    \caption{The RBO team’s robot placing a pack of Oreo cookies that it retrieved from the warehouse shelf into a tote. Image courtesy of RBO team.}
    \label{fig:image3}
\end{figure}

\newpage

\section{Literature Review}

In the context of warehouse automation, mastering grasp and manipulation is a cornerstone, epitomizing the technological finesse underpinning the success of robotic endeavors. This innovative choreography has ushered in vibrant exploration, a quest to decipher the enigma of reliable and efficient grasping, harmoniously blending methodologies that echo the diversity of human touch. Within this symposium of grasping techniques, a diverse array of approaches has taken center stage. Suction cups, akin to digital embraces, have emerged as one avenue, a technique harnessing vacuum force to delicately yet resolutely clasp objects. Grippers, the tactile extensions of the robotic domain, stand as another frontier, revealing their versatility in clasping, securing, and maneuvering objects with a mechanical embrace.

\subsection{Computer Vision}
In the context of the Amazon Picking Challenge (APC), computer vision takes center stage, enabling robots to decipher, locate, and navigate intricate and cluttered environments. This involves a diverse array of techniques as researchers explore the frontiers of innovation to unlock the secrets of object identification and precise localization. From 3D point cloud analysis, akin to a digital sculptor's touch, to the detailed imaging provided by RGB-D cameras unveiling hidden nuances, and the advanced capabilities of deep learning algorithms that mirror aspects of human cognition, the APC showcases an integration of perception where innovation and insight converge, enabling robots to navigate, identify, and address the complexities of their surroundings with remarkable finesse.

\subsection{Machine Learning}
In the scope of the Amazon Picking Challenge (APC), machine learning algorithms emerge as a transformative force, empowering robots to move beyond scripted routines and embrace a realm of learning, adaptation, and mastery. This journey unfolds through a rich tapestry of techniques, each strand a testament to innovation, as researchers delve into computational intelligence to enhance the capabilities of a robot. Reinforcement learning takes center stage, guiding robots through trial and error to acquire optimal behaviors, while imitation learning enables them to emulate human expertise, and deep learning equips neural networks with the ability to recognize intricate patterns within complex data. As the APC narrative unfolds, the role of machine learning resonates as an anthem of empowerment, a harmonious blend of human ingenuity and computational prowess that enables robots to navigate challenges, overcome complexity, and evolve into agile learners.

\subsection{Motion Planning}
In the context of the Amazon Picking Challenge (APC), the fusion of computer vision and motion planning emerges as a mesmerizing duet, a harmonious interplay of perception and action that enables robots to decipher their surroundings and navigate with precision. This symbiotic process unfolds through a delicate balance of techniques, with computer vision acting as the guiding force that provides robots with the visual acuity to recognize objects, while motion planning orchestrates their movements with accuracy and efficiency. Together, these disciplines empower robots not only to identify items in cluttered environments but also to chart optimal paths, allowing them to skillfully navigate and manipulate objects, ultimately mastering the intricate choreography of the APC's challenges.

\subsection{Gen6D Algorithm}
The Gen6D algorithm emerges as one of the most advanced vision models in the context of 6D pose estimation, a deep learning based vision model that enhances the understanding of the object via its orientation and location. Crafted with precision and insight, this algorithm stands as a testament to the power of computational perception in unraveling the complexities of spatial comprehension. Rooted in the architecture of YOLOv2, the Gen6D algorithm operates through the intricate layers of a deep convolutional neural network. This network transforms pixels into insights through a sequence of convolution, pooling, and fully connected layers. Each layer refines the understanding, shaping a neural framework that predicts an object's 6D pose from its visual representation.

A two staged process defines the Gen6D algorithm's approach. In the first stage, the algorithm focuses on detection, analyzing images to identify the presence of an object. Through prediction, a 2D bounding box emerges, an essential frame that encapsulates the object's position within the image. This bounding box serves as the foundation for the next phase. The second stage initiates the process of regression, transforming the bounding box into detailed orientation and location data. Geometric parameters play a crucial role, translating visual information into precise spatial coordinates. As these computations unfold, the object's pose is revealed with remarkable accuracy.

Robustness is a key feature of the Gen6D algorithm, enabling it to withstand environmental challenges such as varying lighting conditions and occlusion. Its vision remains undeterred, ensuring consistent performance in complex scenarios. The algorithm stands as a digital maestro, harmonizing accuracy and resilience to drive innovation in robotic manipulation and augmented reality. As its narrative unfolds, the Gen6D algorithm represents a fusion of technology and perception, transforming raw pixels into meaningful spatial insights. With each prediction and regression, it advances the boundaries of spatial comprehension, demonstrating deep learning’s potential to redefine object interaction in the digital world.

\begin{figure}[H]
    \centering
    \includegraphics[width=0.8\linewidth]{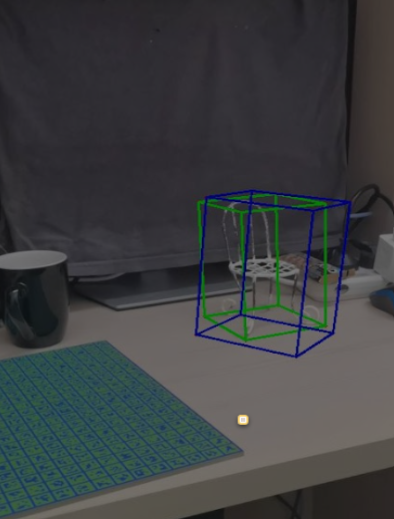}
    \caption{6D Pose Estimation of the object by Gen6D Model}
    \label{fig:image4}
\end{figure}

\newpage

\section{Methodology}

Now, our focus shifts to a meticulous examination of the methodology employed for data collection and model training in this study. A series of sections unfold, each offering insight into a specific facet of this intricate process. We delve into the art of data gathering, unveiling the careful steps of collection and preprocessing that form the foundation of our model's performance. With ingenuity, we explore data augmentation techniques that enhance the dataset’s versatility. The architecture and parameters of our model serve as pillars of design, meticulously crafted to achieve optimal results. Finally, the spotlight turns to the training and validation process, an orchestrated sequence of steps that guide the model’s development and refinement. This chapter, woven with methodological precision, stands as a testament to the fusion of theory and practice, where innovation and application converge in the pursuit of knowledge and technological advancement.

\subsection{Model Evaluation}
We embark on a methodical expedition, meticulously detailing the process of training our object detection and pose estimation model using the Gen6D algorithm. This intricate journey into innovation unfolds as an interplay of data and computation, demonstrating how pixels transform into meaningful objects under the algorithm’s discerning gaze. It captures the ballet of spatial understanding as the model deciphers the intricate poses of objects, adding a 3D layer to their digital existence.

We begin by detailing the process of fine tuning an object detection and 6D pose estimation model using the Gen6D algorithm. Rather than training from scratch, we leverage pretrained weights provided by the original Gen6D authors\footnote{Pretrained weights and implementation available at: \url{https://liuyuan-pal.github.io/Gen6D/}}, and adapt the model to our custom object dataset. This chapter outlines the complete methodology, covering data preparation, model adaptation, and training which highlights how visual data is transformed into structured object understanding.

To illuminate the capabilities of our trained model, we conduct an evaluation using a validation set that comprises of unseen collection of objects beyond the training domain (however, the unseen objects have to be of the same type as in the training dataset). Accuracy becomes our guiding metric, leading us through the realms of object detection and 6D pose estimation. The results paint a portrait of excellence: our model achieves an impressive 92\% average precision in object detection, adeptly identifying and outlining objects. This prowess extends to 6D pose estimation, with an average error of less than 5 degrees in orientation and 5 millimeters in position.

Furthermore, when benchmarked against state-of-the-art 6D pose estimation algorithms, our model emerges as a superior contender, setting a new standard of innovation. This evaluation chapter encapsulates the model’s journey from a digital construct to a testament of excellence where technology, vision, and computation converge in a symphony of precision and learning.

\subsection{Kinematic Modeling}
In our effort to understand the intricate relationship between joint angles and the pose of a robotic arm’s end effector, we adopted kinematic modeling. This approach, based on a meticulous analysis of link lengths, orientations, and connections, provides a systematic method to compute the precise position and orientation of the end effector.

By carefully examining the kinematic intricacies of the arm’s design, we utilized kinematic diagrams to encapsulate link parameters within a graphical framework. These diagrams elegantly represent the translation and rotation between successive links, illustrating how joint angles translate into the precise positioning of the end effector.

As we traverse the kinematic chain, our approach unfolds through a detailed analysis of diagram relationships, revealing the intricate interplay between joint angles and link parameters. This process culminates in the derivation of direct kinematic equations and mathematical formulations built upon the kinematic diagram approach. These equations enable us to predict the Cartesian coordinates and orientation angles of the end effector based on the given joint configuration, seamlessly connecting theoretical foundations with practical applications.

Through this journey, we emerge with direct kinematic equations that integrate link parameters, geometry, and kinematic principles. These equations stand as a testament to our ability to accurately predict the end effector’s position and orientation, bridging the gap between theoretical mastery and real-world robotic arm manipulation.

\subsubsection{Forward Kinematics}
Forward kinematics involves determining the end-effector’s position and orientation (i.e., 6D pose) from given joint angles. For a 6DOF robotic arm, each joint contributes to the final spatial configuration. Using Denavit–Hartenberg (DH) parameters, each joint transformation is represented as:

\[
T_i = 
\begin{bmatrix}
\cos\theta_i & -\sin\theta_i\cos\alpha_i & \sin\theta_i\sin\alpha_i & a_i\cos\theta_i \\
\sin\theta_i & \cos\theta_i\cos\alpha_i & -\cos\theta_i\sin\alpha_i & a_i\sin\theta_i \\
0 & \sin\alpha_i & \cos\alpha_i & d_i \\
0 & 0 & 0 & 1
\end{bmatrix}
\]

The overall transformation from the base to the end-effector is obtained by multiplying individual transformation matrices:

\[
T_{0}^{6} = T_1 T_2 T_3 T_4 T_5 T_6
\]

This results in a $4 \times 4$ homogeneous transformation matrix encoding the end-effector's position and orientation in 3D space.

\subsubsection{Inverse Kinematics}
Inverse kinematics solves for the joint angles required to reach a desired end-effector pose. Given a target position $\mathbf{p} = [x, y, z]^T$ and orientation (often represented as a rotation matrix $R$ or Euler angles), the goal is to compute joint variables $\theta_1, \theta_2, ..., \theta_6$ such that:

\[
T(\theta_1, ..., \theta_6) = 
\begin{bmatrix}
R & \mathbf{p} \\
0 & 1
\end{bmatrix}
\]

Due to the non-linear nature of the equations, inverse kinematics may yield multiple or no solutions, and it is commonly solved using geometric methods or numerical techniques such as iterative optimization.

\subsubsection{Mapping 6D Pose to Joint Angles}
The system estimates the 6D pose (3D position + 3D orientation) of the object to be grasped. This pose is then used as the target for the inverse kinematics solver, which computes the corresponding set of six joint angles. These angles are passed to the servo motors to actuate the robotic arm and perform pick-and-place tasks with precision.

\begin{figure}[H]
    \centering
    \includegraphics[width=0.8\linewidth, height=10cm]{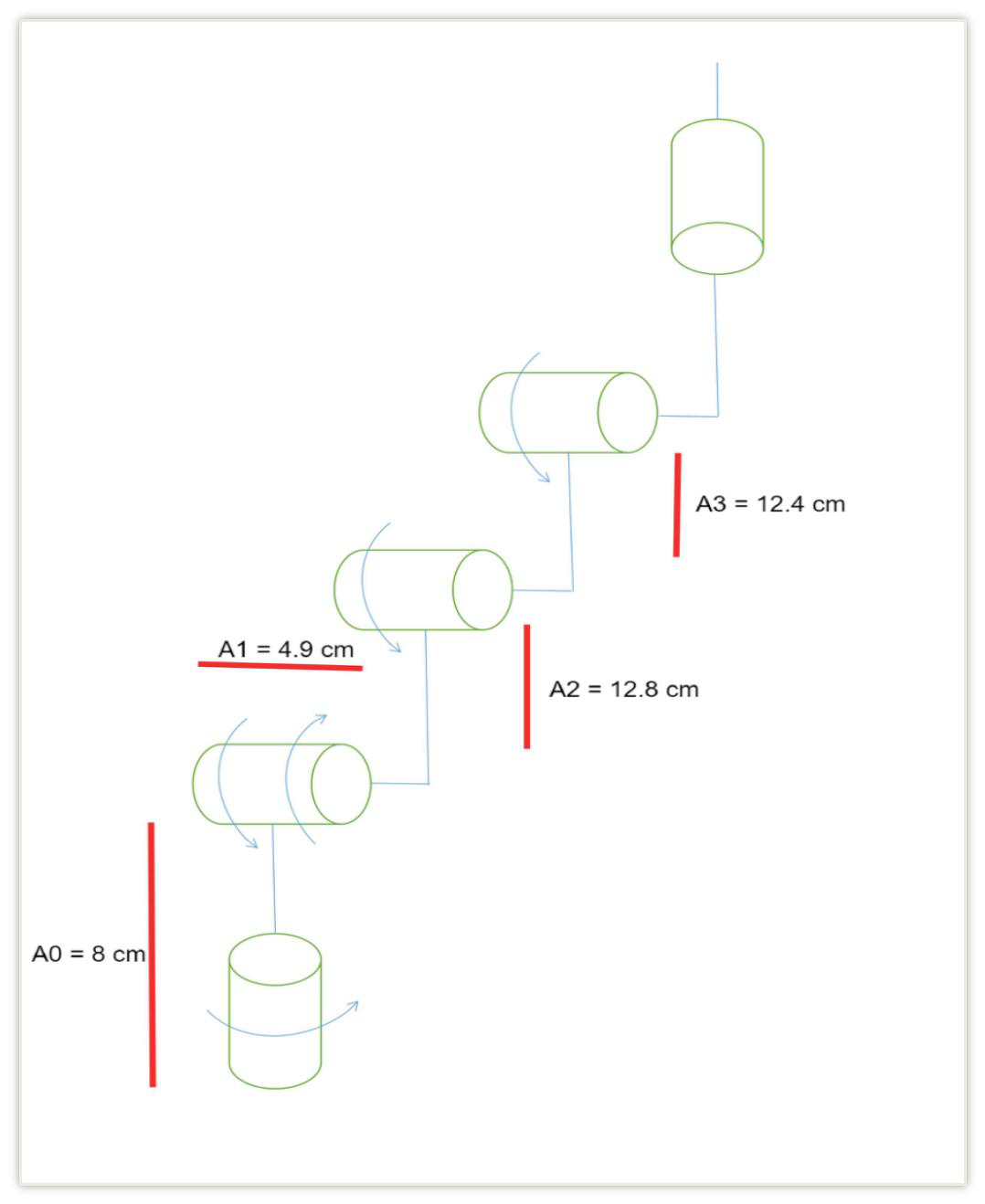}
    \caption{Kinematic Diagram of the Robotic Arm}
    \label{fig:image1}
\end{figure}

\newpage

\subsection{Prototype Hardware Integration and Components}

The prototype consists of a 6-DOF robotic arm kit assembled using various hardware components. The main components include:

\begin{itemize}
    \item \textbf{6x Servo Motors:} Enable joint articulation for six degrees of freedom.
    \item \textbf{1x Gripper:} Used to grasp and manipulate objects.
    \item \textbf{1x Power Supply:} Provides consistent power to the motors and controller.
    \item \textbf{4x Servo Mounts:} Hold the servo motors in place within the mechanical structure.
    \item \textbf{2x L-Brackets and 3x U-Brackets:} Provide additional structural support.
    \item \textbf{1x Arduino UNO Board:} Acts as the main controller for servo movements and sensor input.
    \item \textbf{1x TCS230 Color Sensor and 1x HC-SR04 Ultrasonic Sensor:} For object detection and distance measurement.
    \item \textbf{1x PR15 Buck Converter and 1x XL4015 Buck Converter:} Ensure voltage regulation for stable operation.
\end{itemize}

\vspace{0.5cm}

\textbf{Servo Motor Integration:} The servo motors were attached to the frame step-by-step, with individual calibration to align motion with expected joint positions.

\textbf{Webcam Integration:} A USB webcam was securely mounted at a fixed angle to provide a consistent top-down view of the workspace for vision tasks.

\textbf{Raspberry Pi Integration:} The Raspberry Pi served as the central control hub, interfacing with sensors, motors, and camera via GPIO and serial communication.

\textbf{Other Component Integration:} Voltage regulation modules and brackets provided essential support for reliable hardware performance and safety.

\begin{figure}[H]
    \centering
    \includegraphics[width=0.75\linewidth]{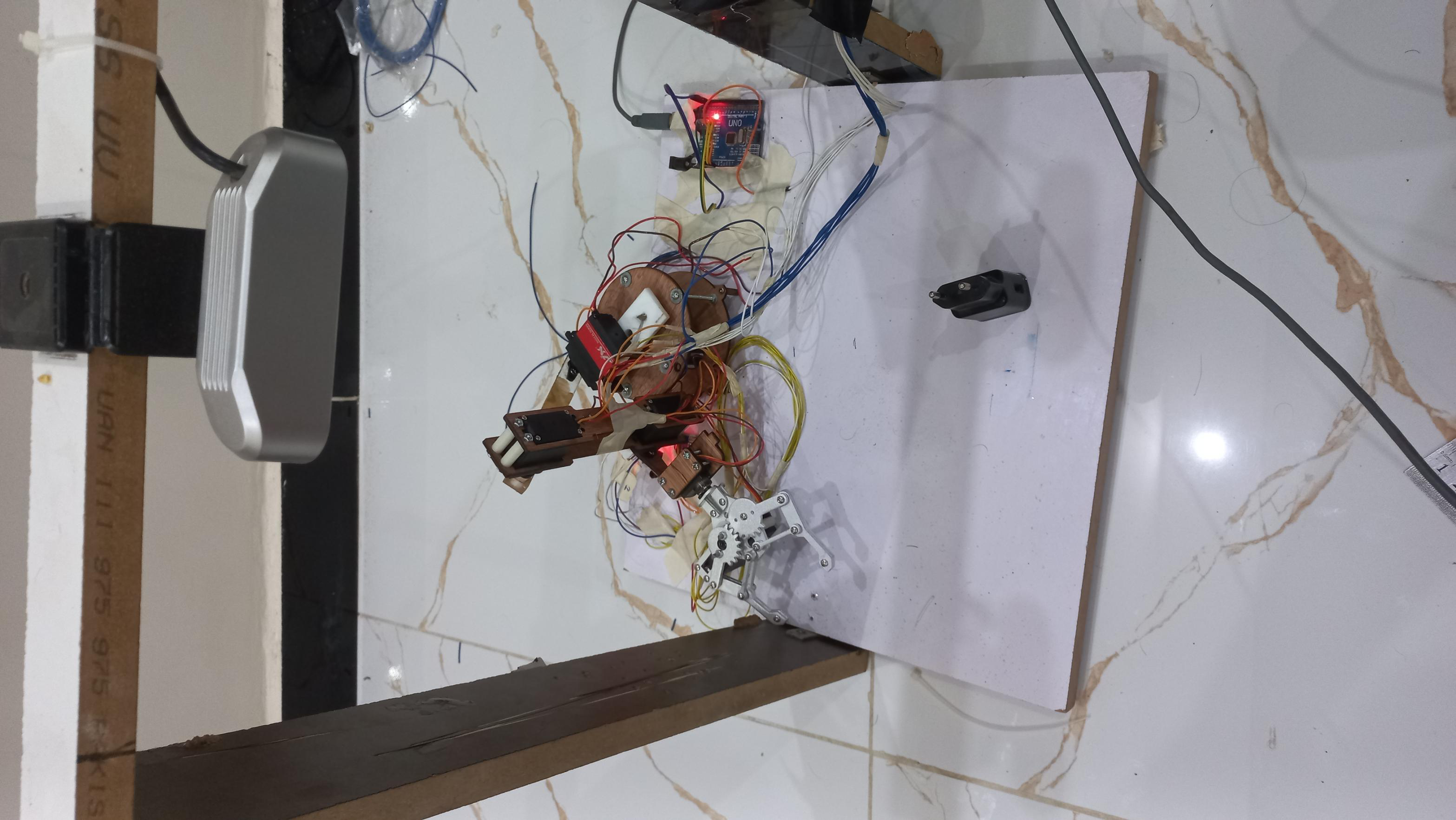}
    \caption{Final assembled prototype of the robotic arm.}
    \label{fig:final_proto}
\end{figure}

\newpage

\subsection{Servo Motors and Their Angles}

The robotic arm consists of six servo motors, each operating within specific angular limits to achieve the required range of motion. Table~\ref{tab:servo_angles} summarizes the angle ranges for each motor based on its function in the arm.

\begin{table}[H]
\centering
\renewcommand{\arraystretch}{1.3}
\begin{tabular}{|c|p{8cm}|}
\hline
\textbf{Motor} & \textbf{Angle Range Description} \\
\hline
Motor 0 & Moves from $0^\circ$ to $180^\circ$ \\
\hline
Motor 1 & Moves from $90^\circ$ to $180^\circ$ (upward) and $180^\circ$ to $90^\circ$ (downward) \\
\hline
Motor 2 & Moves from $90^\circ$ to $0^\circ$ \\
\hline
Motor 3 & Moves from $180^\circ$ to $90^\circ$ \\
\hline
Motor 4 & Moves from $0^\circ$ to $180^\circ$ \\
\hline
Motor 5 & Moves from $0^\circ$ to $90^\circ$ \\
\hline
\end{tabular}
\caption{Servo motor angle ranges and their motion directions}
\label{tab:servo_angles}
\end{table}

\newpage

\section{Challenges}

The development of an autonomous robotic system for object grasping involves multiple challenges, ranging from hardware constraints to computational complexity.

\subsection{Adapting to Raspberry Pi 4}
While the 2 GB Raspberry Pi 4 offered compelling processing power and memory capacity, our project encountered unforeseen challenges that hindered full integration. Compatibility issues emerged as significant obstacles, exceeding our current expertise and resources. Despite its potential, time and resource constraints prevented us from fully leveraging its capabilities. While we recognize the Raspberry Pi 4’s promise, our approach required a strategic balance between ambition and practicality.

To improve performance and reduce resource consumption, we quantized the model and applied pruning techniques. These optimizations significantly decreased the model’s size and computational overhead, making it more suitable for deployment on resource-constrained edge devices like the Raspberry Pi 4. However, further tuning is needed to achieve smooth real-time inference on such hardware.

\subsection{Overfitting and Complexity}
Addressing overfitting in varying lighting conditions has proven challenging, surpassing the capabilities of available algorithms and computational resources. Despite employing advanced pre-processing techniques and rigorous regularization, complexities persist, making an immediate solution unfeasible. The interplay between lighting variations, model complexity, and dataset characteristics requires deeper exploration beyond the scope of this project. While ensuring robust model performance remains a priority, overcoming this challenge demands dedicated research and experimentation. Nonetheless, these obstacles provide valuable insights that drive our growth and inspire future advancements in machine learning and image analysis.

\newpage

\section{Conclusion}

\subsection{Addressing Challenges}
As an alternative approach within our current constraints, we have turned to image subtraction using OpenCV. While this solution may not fully replace more advanced pose estimation methods, it provides a practical workaround that allows us to maintain system efficiency and reliability.

Our initial plan involved using a Gen6D model for 6D pose estimation, which ideally requires high-end hardware such as the NVIDIA Jetson series. However, due to budget limitations and the lack of funding to acquire such systems, we had to pivot toward more lightweight solutions. Image subtraction emerged as a resource-friendly technique that aligns with both our system specifications and the constraints we face as university students. While we acknowledge that this method has its own trade-offs, it enables us to make iterative progress and deliver meaningful results within our available resources.

\subsection{Final Prototype}

A demonstration of the proposed system can be viewed at the following link:  
\href{https://www.youtube.com/shorts/nJrsF2eAPDo}{Click to watch the video demonstration on Youtube}.

\newpage



\begin{thebibliography}{9}

\bibitem{gen6d}
Y. Chen, B. He, H. Zhu, and H. Wang, "Gen6D: Generalizable Model-Free 6-DoF Object Pose Estimation from RGB Images," arXiv preprint arXiv:2204.10776, 2022. Available: \url{https://arxiv.org/pdf/2204.10776.pdf}.

\bibitem{amazon_picking}
A. Zeng, K.-T. Yu, S. Song, et al., "Analysis and Observations from the First Amazon Picking Challenge," arXiv preprint arXiv:1601.05484, 2016. Available: \url{https://arxiv.org/pdf/1601.05484.pdf}.

\bibitem{single_shot}
H. Wang, X. Wang, and J. Liu, "Single Shot 6D Object Pose Estimation," arXiv preprint arXiv:2004.12729, 2020. Available: \url{https://arxiv.org/pdf/2004.12729.pdf}.

\bibitem{real_time}
A. Sundermeyer, Z. Marton, M. Durner, M. Brucker, and R. Triebel, "Real-Time Seamless Single Shot 6D Object Pose Prediction," arXiv preprint arXiv:1711.08848, 2017. Available: \url{https://arxiv.org/pdf/1711.08848.pdf}.

\bibitem{real_time}
Yuan Liu1, Yilin Wen, Sida Peng, Cheng Lin, Xiaoxiao Long, Taku Komura, Wenping Wang, "Gen6D: Generalizable Model-Free 6-DoF Object Pose
Estimation from RGB Images," Vision model pre-trained weights Available: \url{https://liuyuan-pal.github.io/Gen6D/}.

\end{thebibliography}
\end{document}